\documentclass[a4paper]{article}

\usepackage{INTERSPEECH2021}
\usepackage{multirow}
\usepackage{hyperref}

\title{It’s not what you said, it’s how you said it: discriminative perception of speech as a multichannel communication system}
\name{Sarenne Wallbridge, Peter Bell, Catherine Lai}
\address{
  Centre for Speech Technology Research, University of Edinburgh, United Kingdom}
\email{\{s1301730,  peter.bell, c.lai\}@ed.ac.uk}

\begin{document}

\maketitle
\begin{abstract}



People convey information extremely effectively through spoken interaction using multiple channels of information transmission: the lexical channel of \textit{what} is said, and the non-lexical channel of \textit{how} it is said.
We propose studying human perception of spoken communication as a means to better understand how information is encoded across these channels, focusing on the question \emph{What characteristics of communicative context affect listener's expectations of speech?}. 
To investigate this, we present a novel behavioural task testing whether listeners can discriminate between the true utterance in a dialogue and utterances sampled from other contexts with the same lexical content. We characterize how perception -- and subsequent discriminative capability--is affected by different degrees of additional contextual information across both the lexical and non-lexical channel of speech.
Results demonstrate that people can effectively discriminate between different prosodic realisations, that non-lexical context is informative, and that this channel provides more salient information than the lexical channel, highlighting the importance of the non-lexical channel in spoken interaction.





\end{abstract}


\section{Introduction}

In any given communicative context, people make use of all available channels to transmit information as efficiently as possible. Although there is little agreement on how information is encoded across sets of channels, there is no shortage of evidence that communication changes based on the channels provided by the underlying communicative setting \cite{Christiansen2015,Silva2020,Clark2004,Pfau2000,Cohen1984,Jensen2000}. Speech, perhaps the most intrinsic modality of human information transmission, can be modelled as a multi-channel communication system, transmiting information across both  lexical (\textit{what} is said) and non-lexical channels (\textit{how} it is said).

Non-lexical aspects speech are often characterized in terms of prosody, i.e. suprasegmental characteristics of speech such as pitch, loudness, and rhythm, which are usually captured via their acoustic correlates (i.e., fundamental frequency, intensity, duration). Prosody encodes a host of implicit information, from the physical properties of a speaker's vocal tract to their emotional state. However prosodic variation is used explicitly to mark novel information, direct interlocutor attention, and disambiguate lexical information \cite{Ben-David2016,Hirschberg1986,Clark2004,Cervone2018}. Non-lexical language features also have been proposed as tools for acquiring syntactic structure \cite{Pate2013}, as well as mediating the distribution of information transmitted during communication \cite{Aylett2004}.

The value of the non-lexical channel of speech is further exemplified in conversations, which make up the majority of human language use. During interaction, communicative partners engage in an intricate, extremely fast-paced dance of turn management \cite{Pouw2020}. Though the underlying timing mechanisms are highly debated, non-lexical features of speech are widely-accepted as crucial components \cite{Schaffer1983,skantze2020turn}. This use of the non-lexical channel is exploited by infants before language is even learned, and the basic properties are relatively fixed across languages \cite{Levinson2016}.


Although its importance is widely accepted in psycholinguistics, automatic speech representation learning has made little use of the non-lexical channel. Much of the recent work in developing such representations is geared towards downstream tasks such as Automatic Speech Recognition which focus heavily on the lexical content of speech \cite{Chorowski2019, Latif2020}. These representations are designed explicitly to remove information on suprasegmental levels such as emphasis and stress. While non-lexical features are important components of spoken language understanding tasks like automatic emotion recognition or dialogue act detection, these rely on expensive human annotation which condense the rich speech signal into small sets of labels and resulting representations may not generalize well across domains \cite{zhang2017cross}.

We argue that better understanding of how people produce and perceive (i.e., \textit{encode and decode}) information across the transmission channels of speech will enable more effective encoding of contextual information in automatic speech representations. To this end, we present an exploratory study into which characteristics of communicative context, across lexical and non-lexical channels, affect human expectations of spoken communication. We propose a behavioural task, inspired by recent work in self-supervised contrastive learning, to measure the participant's use of lexical and non-lexical channels as a function of their ability to discriminate between utterances with the same lexical content in different contexts.  This discriminative task allows us to investigate how listeners use lexical and non-lexical context to form expectations about dialogue.

\section{Related work} \label{Related work}

To the best of our knowledge, the task presented in this work is the first of its kind to quantify the effect of the non-lexical channel on human perception using discriminative performance and is motivated by findings in numerous related studies. We present some of the most related areas of work below. 

Much of the previous work on non-lexical features and discourse context revolves around learning \emph{spoken discourse structure}. For example, prosodic features can help identify paragraph structure \cite{lai2020integrating}, rhetorical relations \cite{kleinhans2017using}, turn-taking behaviour \cite{gravano2011turn,roddy2018multimodal}, and dialogue acts \cite{shriberg1998can,tran2020neural}.    
These works demonstrate that modelling longer-term non-lexical trends can enhance automatic speech understanding.
However, studies also indicate that lexical and non-lexical content interact in complex ways to signal subtle attitudinal/pragmatic functions \cite{benus2007prosody,lai2010you}. Sparsity of these sorts of functions means that the types of contextual information that can be captured with these supervised learning approaches is limited.  

Those discourse/prosody studies draw on the idea of \emph{discourse coherence}. Coherence characterizes how logical and consistent the internal structure of multi-utterance segment is \cite{Cervone2018}. This complex concept is difficult to operationalize, partially because it attempts capture discourse characteristics as \textit{perceived} by people. Though the concept has primarily been used to assess the quality of texts, 
it has recently been used in both evaluation and objective functions for generative models of written and spoken dialogue.
Modelling of coherence in spoken dialogue has almost exclusively focused on text, however recent studies provide evidence that  additional features such as dialogue acts, some of which may capture prosodic information, outperform purely lexical models \cite{Cervone2020, MesgarSebastian2020}.



Another characteristic of communication that motivates the task we present here is \emph{entrainment}, i.e., the tendency of conversational partners to become more similar to each other. 
This phenomenon has been studied on both lexical and non-lexical dimensions, and has been found to have significant cognitive and social effects, facilitating comprehension and prediction of upcoming speech, as well as perception of rapport, naturalness, and communicative success \cite{Stephens2010,lubold2015naturalness,Levitan2011}. These findings provide evidence that the non-lexical channel carries valuable  information about the semantic/social context. 
Most research in this area has examined behavioural similarity over entire conversations, however studies have shown that alignment changes can produce perceptual effects by conditioning on very localised information: \cite{lubold2015naturalness} finds significant effects on human perception of synthetic dialogue when conditioning on  previous speaker turns. Previous turn features have also been found to be predictive of acoustic properties of upcoming turns \cite{Fuscone2020}. We extend this direction by measuring the effect and utility of varying degrees and types of information on human perception.

Even though local information affects perception, any number of prosodic realisations can still be likely when the context is underspecified \cite{HodariLai2020}. To design a task that captures this complexity of the relationship between communicative channels in speech, we draw inspiration from discriminative training schemes in self-supervised machine learning.  Generally, such schemes involve training a model to discriminate between target and distractor samples. These methods have been extremely successful primarily because they leverage contextual information as a training signal rather than requiring explicit labels \cite{Milde2018}. For our purposes, a discriminative, rather than predictive, task enables us to capture information about the factors that drive expectations of prosody, and to avoid reliance on human-generated labels. 


\section{Experiment Design} \label{Experiment Design}
In the following, we investigate whether, in a given context, listeners can discriminate between turns from that context versus other lexically-equivalent turns sampled from other contexts. 

\subsection{The Discriminative Task} \label{Discriminative Task}
The core of this task involves a dialogue context (T0) and a potential response (T1). 
The task involves presenting participants with the true (T0, T1) sample along with 3 lexically-equivalent (T0, T1) samples where T1 was extracted from elsewhere in the corpus (Section~\ref{Dataset}).
Participants were explicitly instructed that one of the samples was the correct one and then asked to rate each how likely each sample was to be the true one, giving a score from 1 (`Very Unlikely') to 4 (`Very Likely'). 


Along with the transcriptions of T0 and T1, participants were presented with audio stimuli in 2 formats: T1 prosody (text-only context), or T0+T1 prosody (audio+text context). In the T1 format, participants listened to just the responses (T1); in the T0+T1 format, participants were presented with the audio of joined T0+T1 pairs. 
Though pauses can be seen as a joint construction between conversational partners, we modelled pauses as a feature of utterance design as it carries important communicative information \cite{Clark2004,Esposito2008,Ruhlemann2011,Bogels2015,Gorisch2012}. Each response was extracted along with its preceding pause and peak-normalized before concatenation. Pauses could be both positive and negative (i.e., overlapping).




To understand how much context is useful for this task, stimuli in both formats were accompanied by a variable amount of transcript based contextual information for the conversation. Conditions included 0 additional turns (0,0), 3 preceding turns (3,0), 6 preceding turns (6,0), and 3 preceding and 3 proceeding turns (3,3).\footnote{Example audio samples and survey info:  \url{https://data.cstr.ed.ac.uk/sarenne/INTERSPEECH2021/}}


\subsection{Data: The Switchboard Corpus} \label{Dataset}

Experiments were carried out on the Switchboard Telephone Corpus \cite{godfrey1992switchboard} which consists of 
over 2,400 conversations between 542 participants, and includes manual transcriptions and turn segmentations. 
Casual telephone conversations are an ideal data source for this task as speech is spontaneous and transmission is constrained to the speech modality. Compared to other dialogue domains such as interviews, conversational turns are relatively short, providing a large set of lexically-equivalent turns from which to sample.
Potential effects of entrainment based on previous interactions is controlled for by the fact that Switchboard participants don't know each other a priori \cite{Vogels2020}.

To generate the sets of (T0, T1) pairs, the manually-segmented speaker turns were first filtered for some basic inclusion criteria: a response must have a preceding turn containing between 4 and 25 word tokens, and the absolute length of its preceding pause is must be $<$2 seconds. Acceptable response turns were grouped by lexical content, discarding utterances with less than 4 occurrences.  The longest utterance with at least 4 occurrences was 5 tokens. 

50 sets of (T0, T1) pairs were sampled to obtain a uniform distribution of response token lengths (1 to 5 tokens). The experimental set-up thus consisted of 50 sets of context-target (T0, T1) pairs across the 8 conditions described in Section~\ref{Discriminative Task}, resulting in 400 questions in total.

\subsection{The Online Experiment and Participants} \label{Experiment Details}

67 participants were recruited using Prolific Academic, all were native English speakers and based in North America. Participants completed a 20 question Qualtrics survey, taking $\sim$25 minutes to complete. Each participant was presented with one context condition for each sample and participants weren't allowed to complete multiple surveys, ensuring independent responses from different participants for each question.

Each question involved a (T0, T1) discrimination task  (Section~\ref{Discriminative Task}). Additional check questions were interspersed throughout each survey as a basic attention check: select the gender of a speaker in the previous question. Results of participants who obtained less than 70\% overall accuracy on check-questions were rejected.

\subsection{Evaluation Metrics} \label{Evaluation} 
We use accuracy and cross entropy-based metrics to measure the participant performance.
The \emph{accuracy} measure the frequency that the true sample was rated highest, normalised by the number of samples sharing the highest score.
\emph{Cross entropy} measures the deviation between probability distributions $P$ and $Q$. To apply cross entropy to our task, participant ratings were compared to the the ideal scoring--where the correct sample received the maximum score (4) and all other samples, the minimum score (1). This maintained information regarding participants' certainty in the evaluation metric. The 2 score vectors were then converted to probability distributions $P$ and $Q$.


\section{Results} \label{Results}

We report participant performance across stimulus formats and context conditions, along with follow-up analysis on acoustic features and participant certainty.  To assess whether performance is greater than chance, we use two baselines: (i) Accuracy with respect to uniform probability of selecting the true sample (ii) Cross entropy with respect to shuffling each set of human ratings 50 times, i.e., maintaining the distribution of scores. Significance is computed with respect to 95\% confidence intervals around estimated means.

\subsection{Participant Performance}

\begin{figure}[t]
  \centering
  \includegraphics[scale=0.3]{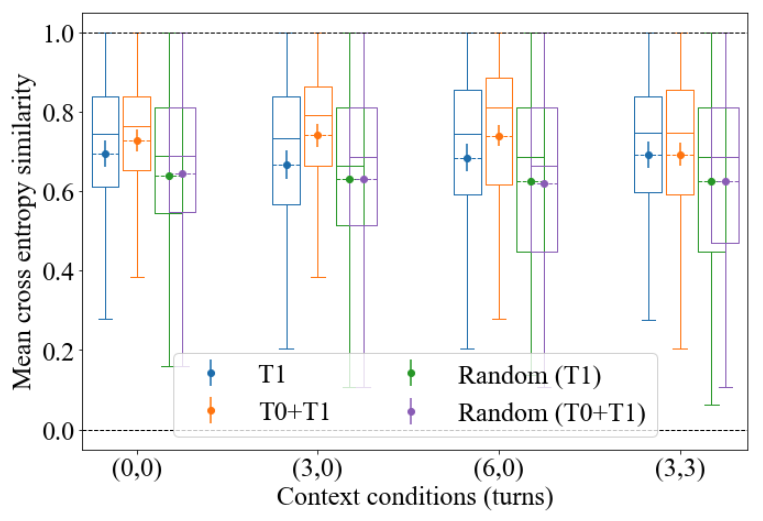}
  \caption{Average cross entropy similarity across context conditions}
  \label{fig:cross_entropy}
\end{figure}

Participant performance is shown in Table \ref{tab:accuracy} (Accuracy) and Figure \ref{fig:cross_entropy} (Cross-Entropy). Performance in both the acoustic+text (T0+T1) and text-only (T1) conditions were significantly better than random across all context conditions. That is, participants were able to discriminate true turns from sampled turns better than chance, and they effectively leveraged non-lexical information for this task.

\begin{table}[t]
 
  \centering
       \caption{Mean participant performance (accuracy). Statistical significance between conditions denoted by *, $\star$, $\bullet$.}
     
  \label{tab:accuracy}
    \begin{tabular}{c|c|c|c|c}
    \hline
        \multirow{2}{*}{\textbf{Prosody format}} &  \multicolumn{4}{c}{\textbf{Context Condition}} \\ 
        \cline{2-5}
        \multirow{2}{*}{} & \textbf{(0,0)} & \textbf{(3,0)} & \textbf{(6,0)} & \textbf{(3,3)} \\
        \hline 
        T1     & 0.346 & \textit{0.329}* & 0.349$^\star$ & \textbf{0.358} \\
        T0+T1  & 0.400 & 0.444* & \textbf{0.468}$^{\star\bullet}$ & \textit{0.378}$^\bullet$ \\
        Random & 0.250 & 0.250 & 0.250 & 0.250 \\
    \hline
    \end{tabular}

  \vspace{-14pt}
\end{table}


With access to response audio (T1) but only text context, participants achieve a mean accuracy of 0.345 across all context conditions, significantly better than if they only had the response text (i.e., all variations equally likely). The significant difference is also reflected in cross-entropy performance. This result provides important evidence for the value of non-lexical information: even with only the transcript of a preceding speaker turn, people can effectively discriminate between different prosodic realisations of the response. 
Interestingly, mean participant performance across context conditions was relatively consistent, showing no significant differences. The fact that additional lexical context didn't affect participant performance provides insight into the value of information contained in the lexical channel (cf. Section \ref{Discussion}). 




Similar to the T1 format, when participants have access to the audio of the immediately preceding turn (T0+T1), accuracy and cross entropy performance is significantly better than the random baselines in the first 3 context conditions, with a mean accuracy of 0.422. This provides further evidence for the importance of the non-lexical channel. 
Performance improves slightly as additional text context is provided (except for the 'future' context condition (3,3)) however, as in the T1 prosody condition, this trend isn't significant. 


Results from both evaluation metrics generally show better performance for T1+T0 compared to T1.  That is, access to audio context helps guide dialogue expectations beyond just having the lexical context. The differences in performance are significant when participants can condition their scores on transcripts of previous speaker turns ((3,0, and (6,0)). However, the difference is not significant for context conditions (0,0) and (3,3). This suggests that beyond a certain point, additional lexical context isn't being leveraged for this task. 

The non-significance in the (0,0) condition could be due to the lack of context available in that condition. However, the fact that participants do significantly worse in condition (3,3) is surprising. This result isn't easily explained by differing effects of past and future contexts.
However, cross entropy performance of participants in condition (3,3) is significantly worse than both (3,0), (6,0). It's possible that this was due to interference: people may struggle to process or integrate such varied information. We leave further investigation for future work. 


\subsection{Acoustic Analysis} \label{Acoustic Analysis}

\begin{figure}[t]
  \centering
  \includegraphics[scale=0.27]{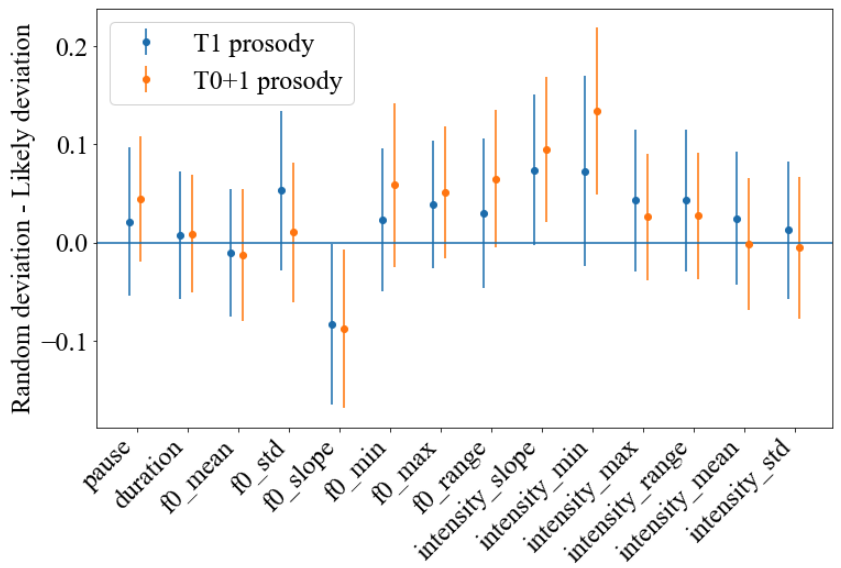}
  \caption{Expected feature deviation in "likely" score sets (across prosody formats). Timing features include pause length before T1, and sample duration}
  \label{fig:diffs}
 \vspace{-14pt}
\end{figure}

The results above indicate that participants  make use of acoustic features when deciding which responses are likely or not. To get a better idea of how listeners utilize this information, 
we analyse some acoustic features commonly used to characterise prosodic in dialogue \cite{Cervone2018,Fuscone2020}. 
Specifically, we would like to know how related similar-scoring samples were in acoustic feature space. 
To measure this, for each stimuli set, we compute the mean deviation between the acoustic representation of each response, i.e., the deviation we would expect if participants scored samples at \emph{random}. 
We then measure the mean deviation between \emph{likely} samples (i.e., rated as either [3, 4]), and we compute the difference between the \emph{random} mean and \emph{likely} mean, scaled by the \emph{random} mean to compare relative differences between features.
Under this metric, a difference of 0 implies that deviations between \emph{likely} samples is not different to randomly rated samples. Differences $>0$ or $<0$ indicate that that \emph{likely} samples are more or less acoustically similar than expected (resp.).

Figure \ref{fig:diffs} shows that, on the whole, samples rated as \emph{likely} tend to be similar in acoustic feature space. However, we don't see significant effects on ratings based on the amounts of acoustic information available (T1 vs T0+T1). Rather, these results highlight that the relationship between score similarity and these prosodic features can't be easily mapped. Similar complexities have been previously reported: \cite{Fuscone2020} finds that acoustic features of speaker turns differ in how much conversational context affected their prediction. Overall, the results indicate that we need to allow for variation in what constitutes `likely' prosody in a specific context, and more flexible ways of describing what that context includes.   


\subsection{Certainty Analysis} \label{Certainty-Analysis}

\begin{table}[t]
      \caption{Distribution of scores across formats and conditions}
  \label{tab:certainty}
  \centering
    \begin{tabular}{c|c|c|c}
        \hline
        \textbf{Prosody} & \textbf{Context} & \textbf{Certain (1,4)} & \textbf{Uncertain (2,3)} \\
        \hline
        \multirow{4}{*}{T1} & (0,0) & 39.22 & 60.78 \\
        & (3,0)                     &  41.19 & 58.81 \\
        & (6,0)                     & 41.13 & 58.87 \\
        & (3,3)                     & \textbf{43.79} & 56.21 \\
        \hline
        \multirow{4}{*}{T0+T1} & (0,0) &  39.08 & 60.92 \\
        & (3,0)                       &  43.79 & 56.21 \\
        & (6,0)                       & \textbf{44.57} & 55.43 \\
        & (3,3)                       &  43.67 & 56.33 \\
        \hline
    \end{tabular}

 \vspace{-16pt}
\end{table}

Though we saw some increases in performance when adding text context to the audio context (T0+T1), the differences were not significant.  
To probe if participants did make use of the extra context, we examined the distribution of scores with respect to certainty across context conditions.
The results in Table \ref{tab:certainty} shows slight differences in how certain participants were across conditions and formats. We see that participants became more certain of their scores as additional context was provided in both T1 and T0+T1 conditions, with the trend being more pronounced in the latter. This indicates that participants were making use of additional context, even if the performance across conditions was not significantly different. Interestingly, participants were slightly more certain with text-only future context (T1, (3,3)), even though they performed closer to the random baseline in that condition.  

\section{Discussion} \label{Discussion}

Overall, the results show that non-lexical context helped participants discriminate actual responses from sampled ones.  However, the results of this exploratory study were not entirely conclusive, indicating future avenues for research into the how expectations arise in dialogue. 


The context conditions were designed to investigate whether people make use of additional context information, and whether this type of information is helpful for the discriminative task. Results in both the text-only and audio+text context conditions remained relatively consistent regardless of how much additional textual context was provided. One explanation is that only local contextual information is needed in this task so additional context isn't made use of. However, the certainty analysis (Section~\ref{Certainty-Analysis}) suggests that participants did take extended text contexts in account when making decisions, though it often seemed to lead to incorrect conclusions.
This gap between channel-based \emph{expectations} about a dialogue and what actually happened needs further investigation, but we can note that the discriminative paradigm used here is useful for this as it allows us to collect a richer, more variable view of listener expectations than other prediction based tasks.  

The differences in mean accuracy between the context formats increased as context was added, which could indicate that salient information in the lexical context can only be leveraged when there is enough non-lexical information. More careful control over the information density of the preceding context may also help untangle the contributions from different channels. Similarly, we would want to extend the types of lexical response we consider. To obtain lexically-equivalent responses, we needed to identify relatively frequent utterances. This meant many of our samples are very short backchannel-style turns (e.g. 'yeah', 'really') which are known to admit wide prosodic variation depending on context \cite{benus2007prosody,lai2010you}. 

The fact that we could only include one audio context turn is a limitation of the current study, and participants may have had difficulty integrating information longer contexts were part text, part audio. This may also account for the dip in performance for the future condition (3,3), so further work should consider adding audio turns after the target response to  investigate the utility of future context. Since Switchboard is made up of short conversations from many speakers, we couldn't select negative samples from only the target speaker (hence give the full matching audio context). Further work using a different speaker data or synthesized speech may help identify long term channel contributions more precisely. 
It would also be useful to analyze this data in the light of listener perceptions of rapport/entrainment, to understand how higher level contextual factors affect dialogue expectations. 

\section{Conclusions} \label{Conclusions}

In this work, we investigated how human expectations of communication are affected by different degrees of contextual information across both the lexical and non-lexical channels of speech.
To do so, we presented a novel behavioural task to measure the isolated effects of lexical and non-lexical context on people's ability to discriminate between different prosodic realisations of lexically-equivalent speaker turns in conversation.  
The relationships between the channels of speech are complex and disentangling their effects on perception requires further investigation. However, these results demonstrate the value of the non-lexical speech channel: 
conditioning  on the non-lexical channel of speech increases people's discriminative ability.
Interestingly, people were able to leverage extremely local information to discriminate between prosodic realisations, achieving significant performance gains by conditioning on only the transcript of the preceding turn. This provides evidence for the feasibility of capturing such information in automatic speech representations, e.g. self-supervised learning methods which leverage relatively local context as a training signal.

\bibliographystyle{IEEEtran}

\bibliography{mybib,clai,discrim_turns}


\end{document}